\documentclass[conference]{IEEEtran}
\usepackage{amsmath,amssymb,amsfonts}
\usepackage{algorithmic}
\usepackage{graphicx}
\usepackage{textcomp}
\usepackage{xcolor}
\usepackage{booktabs}
\usepackage{hyperref}
\usepackage{cite}
\usepackage{algorithm}
\usepackage{enumitem}

\newcommand{\Ex}{\mathbb{E}}
\usepackage{color}

\usepackage{microtype}

\newcommand{\rammstein}{\textbf{\texttt{RAmmStein}}}

\begin{document}

\title{\rammstein: \underline{R}egime \underline{A}daptation in \underline{M}ean-reverting \underline{M}arkets with \underline{Stein} thresholds\textsuperscript{*}\\[6pt]
\small Optimal Impulse Control in Concentrated AMMs}

\author{
    \IEEEauthorblockN{Pranay Anchuri}
    \IEEEauthorblockA{Offchain Labs\\
    panchuri@offchainlabs.com}
}

\maketitle
\def\thefootnote{*}\footnotetext{The authors clarify that \rammstein\ bears no relation to the German industrial metal band of a similar name. We simply find that the free-boundary of our HJB-QVI solution is as solid as a stone (Stein).}

\begin{abstract}
Concentrated liquidity provision in decentralized exchanges presents a fundamental \textbf{Impulse Control} problem. Liquidity Providers (LPs) face a non-trivial trade-off between maximizing fee accrual through tight price-range concentration and minimizing the friction costs of rebalancing, including gas fees and swap slippage. Existing methods typically employ heuristic or threshold strategies that fail to account for market dynamics.

This paper formulates liquidity management as an optimal control problem and derives the corresponding Hamilton-Jacobi-Bellman quasi-variational inequality (HJB-QVI). We present an approximate solution \textbf{\rammstein}, a Deep Reinforcement Learning method that incorporates the mean-reversion speed ($\theta$) of an Ornstein-Uhlenbeck process among other features as input to the model. We demonstrate that the agent learns to separate the state space into regions of action and inaction. We further extend the framework with \textbf{RAmmStein-Width}, which jointly optimizes rebalancing timing and position width via a 6-action DDQN.

We evaluate the framework using high-frequency 1Hz Coinbase trade data comprising over 6.8M trades on a realistic environment (\$10M TVL, 1\% default width). Experimental results show that \rammstein\ achieves a net ROI of 1.60\%, the highest among all realistic (non-omniscient) strategies, while greedy strategies lose up to $-8.4\%$ to gas costs. Notably, the agent reduces rebalancing frequency by 85\% compared to greedy rebalancing. RAmmStein-Width discovers extreme parsimony on its own, executing only 9 rebalances and \$40 in gas, and degrades more slowly than all active strategies at elevated gas costs. Our results demonstrate that regime-aware laziness can significantly improve capital efficiency by preserving the returns that would otherwise be eroded by the operational costs.
\end{abstract}


\section{Introduction}

The rise of Decentralized Finance (DeFi) has fundamentally changed how digital assets are exchanged on blockchains. Automated Market Makers (AMMs) pioneered by Uniswap \cite{adams2021uniswap} lie at the center of this transformation. AMMs are a class of smart contracts that facilitate trade among digital assets without using a traditional order book. These markets are defined by a mathematical invariant that remains constant as users trade against it. Uniswap V2 is the first generation of such markets that use a so-called constant product invariant: $x \cdot y = k$, where $x$ and $y$ are the quantities of the assets in the pool. The marginal price offered by the pool is given by the ratio $\frac{x}{y}$. While the design is quite elegant, it suffers from severe capital inefficiency. The capital deployed by liquidity providers is spread across the entire price spectrum to support market making. For stable asset pairs that trade in a narrow price range, a vast majority of the capital deployed in the pool remains unused. Uniswap V3 \cite{adams2021uniswapv3} addressed this limitation by introducing \textit{concentrated liquidity}, which allows LPs to provide liquidity in a price range of their choice. However, a position earns fees only when the price falls within its range. This innovation increased capital efficiency by orders of magnitude compared to Uniswap V2.

However, concentrated liquidity introduces a significant operational challenge for liquidity providers: \textit{range management}. When the price exits the LP's range, the position becomes inactive and does not earn any fees until (a) the price enters the range naturally, or (b) the LP intervenes and rebalances the position. Each rebalancing event has at least three sources of friction:
\begin{enumerate}
    \item Gas costs: Rebalancing requires at least two on-chain transactions, one to exit the position and another to re-enter at the new price.
    \item Swap fees: When the price exits, the portfolio is composed of a single asset. A portion of this asset must be swapped for the other asset to create fee-earning liquidity again. This swap has an associated fee.
    \item Slippage: For large positions, the swap transaction also incurs a slippage cost.
\end{enumerate}
The liquidity provider faces a dilemma: whether to pay the rebalancing costs and hope to recover them via trading fees, or wait for the price to re-enter the range. Often, the cost to maintain a concentrated position exceeds the marginal fee accrued from that concentration. We call this the \textbf{LP Rebalancing Paradox}.

Existing tools and strategies to manage LPs often make suboptimal decisions because they treat all price deviations identically regardless of the underlying market regime. A 1\% price deviation in a strongly trending market is fundamentally different from a 1\% price change in a mean-reverting regime. Consider the following two scenarios:

\begin{itemize}
    \item \textbf{Scenario A:} Price exits an LP range during a momentum-driven breakout. It is unlikely that the price reverts to the range in the short term. In this scenario, rebalancing is economically justified.
    \item \textbf{Scenario B:} Price exits a narrow-range LP during a noise movement around a stable equilibrium. In this scenario, the probability that the price re-enters the range is high. Therefore, paying the rebalancing costs is not justified.
\end{itemize}

Our paper makes the following novel contributions:

\begin{enumerate}
    \item We formalize liquidity management in concentrated AMMs as an impulse control problem. We derive the corresponding Hamilton-Jacobi-Bellman quasi-variational inequality that completely characterizes the optimal rebalancing policy.
    \item We introduce the \textbf{Stein Signal} ($\theta$), the mean-reversion parameter of an Ornstein-Uhlenbeck process, as a prior that captures market regime dynamics.
    \item We develop \rammstein, a deep reinforcement learning method that approximates the solution of the HJB equation and learns a dynamic boundary separating the continuation region (wait) from the exercise region (rebalance) in the state space.
    \item We conduct extensive backtesting on real-world high-frequency data from Coinbase with realistic pool parameters (\$10M TVL, 1\% width), demonstrating that \rammstein\ achieves the highest net ROI among all realistic active strategies.
    \item We extend the framework to joint timing and width optimization (\textbf{RAmmStein-Width}), showing that our approach generalizes to richer action spaces with regime-dependent width selection and graceful degradation under elevated gas costs.
\end{enumerate}

\section{Background and Related Work}

\subsection{Automated Market Makers}

Automated Market Makers (AMMs) are a class of decentralized exchange protocols that replace traditional order books with deterministic pricing functions implemented as smart contracts. The most prevalent design paradigm is the Constant Function Market Maker (CFMM), in which the reserves of two assets $(x, y)$ must satisfy an invariant $\varphi(x, y) = k$ after every trade. Uniswap V2 popularized the constant product invariant $x \cdot y = k$, distributing liquidity uniformly across all prices from zero to infinity \cite{adams2021uniswap}. Alternative invariants have been proposed to address specific use cases: StableSwap \cite{egorov2019stableswap} employs a hybrid invariant that interpolates between constant-product and constant-sum curves, achieving lower slippage for correlated asset pairs, while Orbital \cite{orbital2024} introduces a geometry-aware invariant adapted to anticipated price trajectories.

A fundamental limitation shared by these constant-function designs is that liquidity is allocated globally across the entire price domain. At any given market price, only a narrow band of the deposited reserves actively facilitates trades; the remainder sits idle, earning no fees while still bearing exposure to adverse price movements. This capital inefficiency is particularly acute for asset pairs that trade within a confined range. Uniswap V3 addressed this limitation through the introduction of \textit{concentrated liquidity}, which we describe next.

\subsection{Concentrated Liquidity Mechanics}

In Uniswap V3, price ranges have boundaries called ticks. The $i$-th tick corresponds to the price $1.0001^{i}$. An LP specifies a lower tick $i_l$ and an upper tick $i_u$, defining the price bounds $p_a = 1.0001^{i_l}$ and $p_b = 1.0001^{i_u}$. To open a position, equal value of both assets is deposited into the pool. The virtual liquidity $L$ within this range is computed as \cite{elsts2021uniswapv3math}:
\begin{equation}
    L = \frac{\Delta x}{\frac{1}{\sqrt{p}} - \frac{1}{\sqrt{p_b}}} = \frac{\Delta y}{\sqrt{p} - \sqrt{p_a}}
\end{equation}
where $\Delta x$ and $\Delta y$ are the deposited token quantities and $p$ is the current price.

The fee accrued by an LP's position during a time period $T$ is given by
\begin{equation}
    \text{Fee} = V_T \cdot \phi \cdot \frac{L_{LP}}{L_{total}}
\end{equation}
where $\phi$ is the pool's fee tier (e.g., 0.05\%, 0.30\%, or 1.00\%), $V_T$ is the total trading volume when the price is within the LP's range.

Critically, when $P \notin [p_a, p_b]$, the position earns zero fees. The LP's capital is fully converted to a single asset (the less valuable one), and the position remains inactive until the price re-enters the range.

\subsection{Impermanent Loss and Rebalancing}

Impermanent Loss (IL) refers to the opportunity cost of providing liquidity versus simply holding the underlying assets. Consider the scenario where an LP opens a position when the price is $p_{init}$ and sets a range $p_a < p_{init} < p_b$. During the course of trading, the price in the pool moves by $\delta$. As the price shifts, the AMM continuously rebalances the LP's portfolio: the average execution price at which the LP effectively sells asset $A$ for asset $B$ lies between $p_{init}$ and $p_{init} + \delta$. The opportunity cost of providing liquidity is then the difference between the value of this rebalanced portfolio and the value of simply holding the original assets at the new price.

For concentrated liquidity positions, this opportunity cost is amplified. Within the active range, higher capital efficiency yields a larger share of trading fees, but once the price exits the range, the position is fully converted to a single asset and ceases to earn fees entirely. Importantly, this opportunity cost arises regardless of the direction of the price movement: whether $\delta$ is positive or negative, the IL depends on the magnitude $|\delta|$ of the deviation from the initial price.

These dynamics create a fundamentally complex optimization landscape for managing a liquidity position. The LP faces the following trilemma.

\begin{itemize}
    \item \textbf{Maximizing in-range time:} The LP earns fees only when the pool price lies within the position's range. Narrow price ranges improve capital efficiency but are breached by smaller price movements.
    \item \textbf{Minimizing rebalancing frequency:} Each rebalancing event incurs multiple sources of friction, including explicit costs such as gas fees and swap fees, as well as implicit costs such as slippage and information leakage.
    \item \textbf{Managing IL exposure:} Aggressive rebalancing crystallizes IL by repeatedly selling low and buying high. Conversely, failure to rebalance leaves capital idle and may further increase the accumulated IL.
\end{itemize}

Previous work has analyzed IL in concentrated liquidity settings \cite{loesch2021impermanent, lambert2022uniswap, milionis2022automated}, deriving closed-form expressions for expected losses under various price path assumptions. However, these analyses typically assume passive position management in which the LP establishes a position and leaves it unchanged. We take a fundamentally different approach by modeling the rebalancing decision as an impulse control problem, where the LP must at each moment decide whether to \textit{continue} with the current position or \textit{jump} to a new recentered position. This formulation connects LP management to the rich literature on impulse control in stochastic systems.

\subsection{Related Work on Liquidity Provisioning}

Several recent works have examined the challenges involved in liquidity provisioning in AMMs. We classify the related work in this area across three themes: a) empirical analyses of LP behavior, b) theoretical frameworks, and c) computational methods.

\subsubsection{Empirical Studies}
Heimbach et al.~\cite{heimbach2021behavior} conducted one of the first empirical analyses of LP behavior in the Uniswap V2 AMM. Their analysis showed that most liquidity providers follow passive strategies and that liquidity provisioning is concentrated in a few pools. Moreover, the impermanent loss dominates the fee revenue generated in pools with volatile asset pairs.
 Berg et al.~\cite{berg2021empirical} analyzed Uniswap and SushiSwap and showed that nearly 30\% of trades on these exchanges happen at suboptimal rates.
 With the introduction of concentrated liquidity in Uniswap V3, Heimbach et al.~\cite{heimbach2022risks} showed that LP returns vary widely depending on position width, price volatility, and fee tier. In addition, profitable positions in volatile pools require active management to overcome impermanent losses.
 Fritsch~\cite{fritsch2021concentrated} evaluated fixed-range and resetting-interval strategies on Uniswap V3 and found that active strategies can outperform passive strategies. In general, their analyses showed that selecting the right strategy is a non-trivial problem.

\subsubsection{Theoretical Frameworks}
Evans~\cite{evans2020liquidity} derived LP share returns for geometric mean market makers (such as Uniswap and Balancer) and showed that LP shares replicate derivative payoffs.
Cartea et al.~\cite{cartea2024predictable} formulated a continuous-time stochastic control problem for strategic LPs. They introduced the concept of predictable loss (PL) and derived a closed-form optimal strategy (under log-utility conditions) that balances PL and the collected fees.
Milionis et al.~\cite{milionis2023myersonian} applied Myerson's optimal auction theory to design incentive-compatible AMMs. In their design, the LP chooses a demand curve that has a bid-ask spread, inducing the traders to submit a true estimate of the asset price to trade in the AMM.
Bar-On and Mansour~\cite{baron2023online} framed LP as an online learning problem and provided lower bounds on LP reward using regret minimization without assumptions on future price distributions.

\subsubsection{Computational Methods}
Fan et al.~\cite{fan2024strategic} introduced $\tau$-reset strategies for Uniswap V3. They showed that dynamic liquidity provisioning strategies optimized via stochastic optimization techniques exhibit higher LP earnings compared to baseline Uniswap V2 allocation.
Urusov et al.~\cite{urusov2025tau} further extended this using an ML ensemble combined with a backtesting framework to incorporate historical market dynamics.
Zhang et al.~\cite{zhang2023adaptive} applied a Dueling DDQN with a Loss-Versus-Rebalancing (LVR) reward function that includes hedging via futures. Their agent accounts for gas fees but does not incorporate mean-reversion signals or formulate the problem as impulse control.
Xu and Brini~\cite{xu2025improving} used Proximal Policy Optimization with an LVR penalty and a rolling window training approach, outperforming passive LP in the majority of evaluation windows.
Jaimungal et al.~\cite{jaimungal2023optimal} solved an HJB equation for optimal AMM trading using the Deep Galerkin Method, but focused on execution strategy rather than LP management.

Our work differs from these approaches in three key ways. First, we formalize LP management as an impulse control problem and derive the HJB-QVI, connecting our agent's behavior to the optimal control literature.
Second, we introduce the Stein Signal ($\theta$) as a regime indicator that allows the agent to distinguish between temporary noise and long-term trends.
Third, our agent learns a laziness boundary that balances rebalancing costs against the probability of natural price reversion, achieving significant savings by avoiding unnecessary interventions.

\subsection{Impulse Control Theory}

Optimal control theory studies systems whose state evolves according to known dynamics, subject to control inputs applied over time. The objective is typically to choose a control policy that minimizes a cost functional (or equivalently, maximizes a reward) over a given horizon. The underlying system may evolve deterministically or stochastically, and the control may be applied continuously or at discrete intervention times.

Impulse control problems constitute a subclass of optimal control in which the controller makes discrete interventions at self-selected times, with each intervention incurring a fixed cost independent of the system state. Unlike continuous control, where actions are applied smoothly over time, or optimal stopping, where a single terminal decision is made, impulse control permits an arbitrary sequence of instantaneous interventions separated by periods of passive observation. This framework is particularly well-suited to settings where friction costs render frequent interventions prohibitively expensive. This is precisely the situation faced by LPs, for whom each rebalancing event incurs gas fees, swap costs, and slippage.

The seminal work by Bensoussan and Lions \cite{bensoussan1984impulse} established a rigorous mathematical framework for impulse control in diffusion processes. Their key insight was that optimal policies partition the state space into two distinct regions: a \textit{continuation region}, where the controller optimally waits and allows the system to evolve according to its natural dynamics, and a \textit{jump region}, where immediate intervention is optimal. The resulting optimal policy exhibits a threshold structure: the controller waits until the system state reaches the boundary between these regions, then intervenes to reset the state. The location of this boundary depends on the running reward, the intervention cost, and the stochastic dynamics of the underlying process.

For a controlled diffusion, the value function $V(x)$, representing the expected discounted future reward from state $x$ under the optimal policy, satisfies a quasi-variational inequality (QVI). This inequality takes the form of a complementarity condition:
\begin{equation}
    \max\left\{ \mathcal{L}V(x) + f(x), \mathcal{M}V(x) - V(x) \right\} = 0
\end{equation}
where $\mathcal{L}$ denotes the infinitesimal generator of the diffusion, encoding both the drift and volatility of the process; $f(x)$ represents the instantaneous running reward accrued while waiting; and $\mathcal{M}$ is the intervention operator that maps the current value to the post-jump value minus the jump cost. The QVI asserts that at every point in the state space, at least one of two conditions holds with equality. In the continuation region, the first term equals zero, meaning the value function satisfies the Hamilton-Jacobi-Bellman (HJB) equation, the fundamental partial differential equation of stochastic optimal control that characterizes the value of following an optimal policy in the absence of interventions. In the jump region, the second term equals zero, indicating that the current value exactly equals the value attainable through immediate intervention.

\subsection{Mean-Reversion in Market Microstructure}

The Ornstein-Uhlenbeck (OU) process originates from the modeling of physical systems that exhibit a mean-reverting tendency. It was introduced by Uhlenbeck and Ornstein \cite{uhlenbeck1930theory} as a physically motivated alternative to Brownian motion, which models purely diffusive, memoryless dynamics. Since the foundational work of Vasicek \cite{vasicek1977equilibrium} on interest rate modeling, the OU process has been extensively employed to capture mean-reverting behavior in financial markets. While long-horizon price movements in efficient markets are generally modeled as geometric Brownian motion, high-frequency price dynamics often exhibit pronounced mean-reversion arising from the microstructure of trading itself. Several mechanisms generate this behavior naturally:

\begin{itemize}
\item Market maker inventory management: When a market maker accumulates excessive inventory in one direction, they shade their quotes to encourage offsetting flow, creating a restoring force that pulls prices back toward recent averages.
\item Cross-venue arbitrage: When an asset trades on multiple markets and the price dislocates on one venue, arbitrageurs trade against the dislocation, driving the price back toward the consensus level across markets.
\item Order book resilience: Large trades that consume liquidity on one side of the book are often followed by the gradual replenishment of limit orders, creating buying pressure after price declines and selling pressure after price increases.
\end{itemize}

\subsubsection{Stochastic Differential Equations and the OU Process}

A continuous-time stochastic process $X_t$ is said to satisfy a stochastic differential equation (SDE) if it obeys
\begin{equation}
    dX_t = b(X_t)\,dt + \sigma(X_t)\,dW_t, \quad X_0 = x
\end{equation}
or equivalently, in integral form,
\begin{equation}
    X_t = x + \int_0^t b(X_s)\,ds + \int_0^t \sigma(X_s)\,dW_s
\end{equation}
where the second integral is an It\^{o} integral with respect to a standard Wiener process $W_t$. The function $b(\cdot)$ governs the deterministic drift of the system, while $\sigma(\cdot)$ modulates the intensity of random fluctuations.

The OU process is the specific case in which the drift is linear and the diffusion coefficient is constant. It satisfies the SDE
\begin{equation}
    dS_t = \theta(\mu - S_t)\,dt + \sigma\, dW_t
\end{equation}
The drift term $\theta(\mu - S_t)$ is proportional to the deviation from the long-run mean $\mu$, with $\theta > 0$ governing the speed of mean-reversion. The stochastic term $\sigma\, dW_t$ represents random fluctuations that continuously perturb the process away from equilibrium.

The mean-reverting property of the OU process also admits a statistical interpretation through Stein's paradox. Stein \cite{efron1977stein} demonstrated that shrinkage estimators, which implicitly incorporate mean-reversion assumptions, can dramatically improve estimation accuracy relative to maximum likelihood, even in settings where such assumptions appear unjustified a priori.

The parameter $\theta$ is of central importance to our framework. In our context, $\theta$ serves as a real-time regime indicator: high values signal a strongly mean-reverting environment where price deviations are likely temporary, while low values indicate trending or random-walk behavior where deviations tend to persist or amplify. Looking ahead, the central insight underlying our approach is that when $\theta$ is large, it is preferable for the LP to wait for the price to revert naturally rather than incur the cost of rebalancing; conversely, when $\theta$ is small, the price is more likely drifting, and the LP should rebalance promptly or adopt a more conservative position. The characteristic timescale of mean-reversion is given by the half-life $t_{1/2} = \ln(2)/\theta$. For $\theta = 0.01$ per second, the half-life is approximately 69 seconds, meaning that a price deviation from $\mu$ is expected to decay by half within just over one minute. An LP observing such a $\theta$ estimate may rationally defer rebalancing and instead wait for the price to naturally revert into the active range.

\subsection{Deep Reinforcement Learning for Finance}

Deep Reinforcement Learning (DRL) combines the representational power of neural networks with the sequential decision-making framework of reinforcement learning, enabling agents to learn complex policies directly from high-dimensional observations. This combination has achieved remarkable success across diverse domains, including game playing \cite{silver2016mastering} and robotic control \cite{levine2016end}. More recently, DRL has emerged as a prominent methodology in quantitative finance. In financial contexts, DRL has been applied to portfolio optimization \cite{jiang2017deep}, where agents learn to dynamically allocate capital across assets; market making \cite{spooner2018market}, where agents learn quote placement strategies that balance inventory risk against profit opportunities; optimal execution \cite{ning2021double}, where agents learn to split large orders across time to minimize market impact; and optimal batch posting \cite{mamageishvili2023batch}, where agents learn to time data batch submissions to minimize posting fees. These applications share a common structure: the agent observes market features, takes actions that affect positions or orders, and receives rewards based on realized profits or costs.

In Reinforcement Learning (RL) \cite{sutton2018reinforcement}, the primary goal is to learn a value function that measures the expected discounted cumulative reward attainable by an agent. Traditional methods, such as tabular value iteration, require repeated sweeps of value and policy improvement to converge. This approach becomes intractable in high-dimensional state spaces, where the number of states grows exponentially with the number of features. In DRL, a deep neural network serves as a function approximator that learns compact representations of these complex state spaces, enabling generalization across similar states.

Building on this, the Deep Q-Network (DQN) architecture introduced by Mnih et al.\ \cite{mnih2015human} represents a foundational approach to value-based DRL. The key insight is to approximate the action-value function $Q(s, a)$, which represents the expected cumulative discounted reward from taking action $a$ in state $s$ and thereafter following the optimal policy, using a neural network with parameters $\theta$. The network is trained via temporal difference learning \cite{sutton2018reinforcement}: given a transition $(s, a, r, s')$, the loss penalizes the squared difference between the current estimate $Q(s, a; \theta)$ and the target $r + \gamma \max_{a'} Q(s', a'; \theta)$, where $\gamma$ is the discount factor. DQN introduces two critical mechanisms to stabilize training: (i) experience replay, which stores transitions in a buffer and samples mini-batches uniformly to break temporal correlations; and (ii) a target network, a slowly updated copy of the Q-network used to compute stable target values.

The Double DQN variant, introduced by van Hasselt et al.\ \cite{van2016deep}, addresses an important limitation of standard DQN. Because the same network is used both to select the maximizing action and to evaluate its value, the max operator introduces a systematic upward bias: the maximum of noisy estimates exceeds the true maximum in expectation, leading to overestimation of Q-values. Double DQN decouples these two functions by using the online network for action selection and the target network for value evaluation:
\begin{equation}
    y = r + \gamma Q(s', \arg\max_{a'} Q(s', a'; \theta); \theta^-)
\end{equation}
Here $\theta$ denotes the online network parameters and $\theta^-$ denotes the target network parameters. This simple modification substantially reduces overestimation bias and improves learning stability.

Our work applies Double DQN to the impulse control problem of LP rebalancing. Although the action space is binary (wait versus rebalance), the optimal policy depends on a complex interplay of factors including price dynamics, regime indicators, and position status. The neural network learns to approximate the boundary between the continuation and jump regions without requiring an explicit solution to the HJB equation, instead discovering this boundary through trial-and-error interaction with a simulated environment.

\section{Theoretical Framework}
\label{sec:theory}

In this section, we describe the theoretical framework used to model the problem of managing a liquidity position in a Uniswap V3-style AMM.

\subsection{Problem Statement}
Consider a liquidity provider with capital $K$ deployed in a concentrated position in a Uniswap V3 pool. Instead of using lower and upper ticks, we characterize the position by a center price $c$ and a total width of $2w$. The position earns fees only when the price $S_t$ lies within the interval $[c(1-w), c(1+w)]$. When $S_t$ exits this range, the position becomes entirely single-asset, depending on the direction in which the price exits.

The liquidity provider faces a sequential decision problem over a time horizon $[0, T]$. At each instant $t$, the provider can choose to either allow the position to continue in its current state or pay a cost $C$ to recenter it at the current price $S_t$. The cost $C$ encompasses the gas fee for the blockchain transaction, swap fees for rebalancing the token composition, and slippage. The provider's objective is to choose a sequence of intervention times that maximizes the expected cumulative discounted profit:

\begin{equation}
\label{eq:objective}
    J = \Ex\left[ \int_0^T e^{-\rho t} f(S_t, c_t) dt - \sum_{i} e^{-\rho \tau_i} C \right]
\end{equation}

The first term in the objective function represents the cumulative fee income discounted to present value at rate $\rho$. The instantaneous fee rate $f(S_t, c_t)$ is proportional to trading volume when the position is active and zero otherwise:
\begin{equation}
    f(S_t, c_t) = \begin{cases} V_t \cdot \phi \cdot \dfrac{L_{LP}}{L_{total}} & \text{if } S_t \in [c_t(1-w),\; c_t(1+w)] \\[6pt] 0 & \text{otherwise} \end{cases}
\end{equation}
where $V_t$ is the instantaneous trading volume, $\phi$ is the pool fee tier, $L_{LP}$ is the LP's liquidity, and $L_{total}$ is the total pool liquidity.

The second term represents the cumulative intervention costs over the entire horizon, discounted at rate $\rho$. We assume that each rebalancing event incurs a fixed cost $C$.

The expectation in Equation~\ref{eq:objective} is taken over the stochastic price path, and the optimization is over all admissible sequences of intervention times.

This formulation is an instance of the impulse control problems discussed in Section~II-D, where the goal is to select discrete intervention times that maximize net reward. As noted therein, the optimal policy for such problems exhibits a threshold structure: there exists a boundary in the state space separating the \textit{continuation region}, where the provider optimally waits, from the \textit{jump region}, where immediate intervention is optimal.

\subsection{The Price Process Model}

We model the mid-price $S_t$ as an Ornstein-Uhlenbeck process (see Section~II-E for background):
\begin{equation}
    dS_t = \theta(\mu - S_t)\,dt + \sigma\, dW_t
\end{equation}

To obtain the closed-form solution, define $g(t) = S_t\, e^{\theta t}$. Applying It\^{o}'s lemma:
\begin{equation}
    dg = e^{\theta t}\left(\theta S_t\, dt + dS_t\right) = e^{\theta t}\left(\theta \mu\, dt + \sigma\, dW_t\right)
\end{equation}
Integrating both sides from $0$ to $t$ and rearranging yields the explicit solution:
\begin{equation}
\label{eq:ou_solution}
    S_t = \mu + (S_0 - \mu)\,e^{-\theta t} + \sigma \int_0^t e^{-\theta(t-s)}\, dW_s
\end{equation}
The first two terms describe deterministic relaxation toward the mean $\mu$ at rate $\theta$, while the stochastic integral captures the accumulated effect of random perturbations, each exponentially discounted by the time elapsed since its occurrence.

The parameters $(\theta, \mu, \sigma)$ are estimated via rolling OLS regression on a sliding window of prices. Given discrete prices $\{S_1, \ldots, S_n\}$, we fit the regression:
\begin{equation}
    S_{t+1} - S_t = \alpha + \beta S_t + \epsilon_t
\end{equation}
The OU parameters are then recovered as $\theta = -\beta/\Delta t$, $\mu = -\alpha/\beta$, and $\sigma$ is estimated from the residual standard deviation.

A high $\theta$ indicates strong mean-reversion (price is likely to return to the mean), while a low $\theta$ indicates trending behavior (price deviations tend to persist).

\subsection{Optimal Control Framework for LP management}

The Hamilton-Jacobi-Bellman (HJB) equation is the central equation in optimal control problems involving continuous-time deterministic and stochastic processes. It is based on Bellman's principle of optimality, which states that for all initial states and decisions, the remaining decisions must constitute an optimal policy with respect to the state resulting from the first decision.

\subsection*{1. Definition of the Value Function}
We define the value function $V(S, c, t)$ as the maximum expected discounted profit from time $t$ to the horizon $T$, given the current asset price $S$ and the current center of the liquidity position $c$:
\begin{equation}
\begin{split}
V(S, c, t) = \sup_{\{\tau_i\}} \mathbb{E} \bigg[ \int_{t}^{T} e^{-\rho(\tau-t)} f(S_\tau, c_\tau) d\tau \\
- \sum_{\tau_i \geq t} e^{-\rho(\tau_i-t)} C \mid S_t=S, c_t=c \bigg]
\end{split}
\end{equation}

where $f$ is the instantaneous fee reward and the asset price $S_t$ follows an OU process:
\begin{equation*}
dS_t = \theta(\mu - S_t)dt + \sigma dW_t
\end{equation*}

\subsection*{2. Derivation of the HJB for the Continuous Regime}
We first define the HJB governing the continuous \textit{continuation regime}, where the liquidity provider chooses to let the position evolve without any intervention. Without intervention, the center $c$ remains constant. Applying Bellman's optimality principle:

\begin{equation}
\begin{split}
V(S, c, t) = \max_{u} \mathbb{E} \bigg[ \int_{t}^{t+\Delta t} e^{-\rho(\tau-t)} f(S_\tau, c) d\tau \\
+ e^{-\rho \Delta t} V(S_{t+\Delta t}, c, t+\Delta t) \bigg]
\end{split}
\end{equation}

For infinitesimal $\Delta t$, the integral is approximated as $f(S, c)\Delta t$. We expand the future value term using It\^{o}'s lemma and the approximation $e^{-\rho \Delta t} \approx 1 - \rho \Delta t$ to obtain
\begin{equation}
\begin{split}
V \approx \mathbb{E} \bigg[ f(S, c)\Delta t + (1 - \rho \Delta t) \bigg( V + \frac{\partial V}{\partial t} \Delta t \\
+ \frac{\partial V}{\partial S} dS + \frac{1}{2} \sigma^2 \frac{\partial^2 V}{\partial S^2} \Delta t \bigg) \bigg]
\end{split}
\end{equation}

Substituting the expectation of the OU drift $\mathbb{E}[dS] = \theta(\mu - S)\Delta t$, dividing by $\Delta t$, and letting $\Delta t \to 0$, we obtain the HJB equation for the value function of the LP position in the continuation region:
\begin{equation}
\rho V - \frac{\partial V}{\partial t} = f(S, c) + \theta(\mu - S) \frac{\partial V}{\partial S} + \frac{1}{2} \sigma^2 \frac{\partial^2 V}{\partial S^2}
\end{equation}

\subsection*{3. The Impulse Control Quasi-Variational Inequality}

In the full sequential decision problem, the provider chooses between two actions: allow the position to evolve naturally or pay a cost $C$ to rebalance the center to the current price. This implies a discrete jump in the state space where the center is reset to the current market price, $c \to S$.

The optimal value function $V(S, c, t)$ must satisfy the following Quasi-Variational Inequality (QVI):
\begin{equation}
\label{eq:hjbqvi}
\min \left[
\begin{aligned}
&\rho V - \frac{\partial V}{\partial t} - \mathcal{L}_{OU}V - f(S, c), \\
&V(S, c, t) - \left(V(S, S, t) - C\right)
\end{aligned}
\right] = 0
\end{equation}
where $\mathcal{L}_{OU}$ is the infinitesimal generator for the OU process:
\begin{equation}
\mathcal{L}_{OU}V = \theta(\mu - S) \frac{\partial V}{\partial S} + \frac{1}{2} \sigma^2 \frac{\partial^2 V}{\partial S^2}
\end{equation}
The first term captures the mean-reverting drift: when the current price $S$ is above the long-run mean $\mu$, the drift is negative, reducing $V$. The second term captures how uncertainty in the price process affects the value function through the curvature of $V$.

This formulation identifies two distinct regions in the state space:
\begin{itemize}
    \item \textbf{Continuation Region}: The first term in Equation~\ref{eq:hjbqvi} equals zero. The value function evolves according to fee accrual and price dynamics. In this region, the marginal value of waiting exceeds the marginal value of acting.
    \item \textbf{Jump Region}: The second term in Equation~\ref{eq:hjbqvi} equals zero. The value of the current state $(S, c)$ is exactly the value of the recentered state $(S, S)$ minus the cost $C$.
\end{itemize}

\subsection{The Laziness Principle}

We define the \textit{Laziness Boundary} as the boundary separating these two regions. Although the Double DQN agent learns to approximate this boundary without explicitly solving the QVI, we can characterize its properties by analyzing the trade-off between rebalancing costs and the strength of mean-reversion.

\subsubsection{Rebalancing trade-off}
The fundamental question facing an LP when the price exits the active range is whether to rebalance (paying cost $C$ to restore fee accrual) or wait for the price to re-enter the range naturally. This decision depends critically on the estimated mean-reversion speed $\theta$. To formalize, consider the value of waiting, $V_{\text{wait}}$, from a state where the position is out of range:

\begin{equation}
V_{\text{wait}}(S_t) = \mathbb{E} \left[ \int_{t}^{\tau} e^{-\rho(\xi-t)} f(S_\xi) d\xi + e^{-\rho(\tau-t)} V(S_\tau) \mid S_t \right]
\end{equation}

where $\tau$ is the first passage time at which the price either returns to the active range or exits to a state where rebalancing becomes optimal. Since $f(S_\xi) = 0$ while out of range, the integral term vanishes and $V_{\text{wait}}$ reduces to the discounted expected value at the first passage time $\tau$.

\subsubsection{First Passage Probabilities}

We now analyze the probability of the process returning to the long-run mean $\mu$ before hitting an outer barrier price $L$. This is given by the ratio of scale functions \cite{elliott2005}:

\begin{equation}
P(\text{return}) = \frac{\int_{L}^{s} \exp\left( \frac{\theta(y-\mu)^2}{\sigma^2} \right) dy}{\int_{L}^{\mu} \exp\left( \frac{\theta(y-\mu)^2}{\sigma^2} \right) dy}
\end{equation}

As the mean-reversion strength $\theta$ increases, $P(\text{return}) \to 1$ even for starting states $s$ significantly far from $\mu$. This confirms that during periods of high $\theta$, price reversion is nearly certain.

 \subsubsection{Option interpretation}
When the price exits the range, the liquidity provider holds an American-style \textit{option to wait}, where $C$ acts as the strike price and $\theta$ governs the option's moneyness. The DDQN agent learns to approximate the optimal exercise boundary by balancing the cost of foregone fee accrual against the strike price of rebalancing.

\subsection{DDQN as an HJB Solver}

Equation~\ref{eq:hjbqvi} provides a complete characterization of the optimal policy, but solving it numerically is challenging: the equation is nonlinear due to the min operator, the state space becomes high-dimensional when OU parameters are included, and the parameters themselves are time-varying and estimated with error.

DRL offers an alternative approach that sidesteps explicit solution of the QVI. The key connection is between the HJB equation governing the continuation region and the Bellman equation that underlies Q-learning. Discretizing time into intervals of length $\Delta t$, the HJB equation in the continuation region can be written as
\begin{equation}
    V(s, c) = f(s, c) \Delta t + e^{-\rho \Delta t} \Ex[V(S_{t+\Delta t}, c)]
\end{equation}
This can be rewritten as the Bellman equation $V(s) = r + \gamma \Ex[V(s')]$, where $\gamma = e^{-\rho \Delta t}$ is the discount factor and $r = f \Delta t$ is the immediate reward.

The Q-function extends this to action-conditioned values. In our setting the Q-values for the two actions, wait ($a=0$) and rebalance ($a=1$) are
\begin{align}
    Q(s, c, a=0) &= f(s, c) \Delta t + \gamma \Ex[V(S', c)] \\
    Q(s, c, a=1) &= -C + \gamma \Ex[V(S', S')]
\end{align}
The value function is then $V(s,c) = \max_a Q(s, c, a)$, and the optimal policy selects the action with higher Q-value. The DDQN algorithm approximates the Q-function using a neural network $Q(s, a; \theta)$, where $\theta$ denotes the network parameters. Given transitions $(s, a, r, s')$ sampled from the replay buffer, the network minimizes:
\begin{equation}
    \mathcal{L}(\theta) = \Ex\left[ \left( Q(s, a; \theta) - y \right)^2 \right]
\end{equation}
where the target $y = r + \gamma Q(s', \arg\max_{a'} Q(s', a'; \theta); \theta^-)$ uses the Double DQN formulation to reduce overestimation bias.

Upon convergence, the learned Q-function implicitly encodes the solution to Equation~\ref{eq:hjbqvi}. The boundary between states where $Q(s, 0) > Q(s, 1)$ and those where $Q(s, 1) > Q(s, 0)$ corresponds to the Laziness Boundary derived from the HJB analysis. The DDQN approach handles nonlinearity through function approximation and naturally accommodates time-varying parameters without requiring closed-form solutions. Notably, the estimated OU parameters are included as input features to the Q-network alongside the state variables, providing the agent with explicit regime information.

\section{Methodology}
In this section, we describe how the theoretical framework of Section~\ref{sec:theory} can be used to build a \rammstein\ system that effectively manages liquidity positions in AMMs. Although the description assumes Uniswap V3-style AMMs, the system is sufficiently general to accommodate AMMs with different characteristics.

\subsection{System Architecture}

The \rammstein\ system consists of three main components:

\begin{enumerate}
    \item \textbf{Feature Engine}: This component is responsible for computing the OU parameters and constructing the state representation.
    \item \textbf{Environment Simulator}: This simulates the dynamics of liquidity positions, fee accrual, and rebalancing costs.
    \item \textbf{DDQN Agent}: Learns a solution for the HJB-QVI equation through experience replay and temporal difference learning.
\end{enumerate}

\subsection{State Representation}

The agent observes an 8-dimensional state vector $\mathbf{s}_t$ at each timestep, capturing the relationship between the current price $S_t$, the position center $c$, and the estimated OU parameters. The components are defined as follows:

\begin{itemize}
    \item \textbf{Normalized Price Deviation} ($\delta_p = S_t/c - 1$): Represents the current deviation of the market price from the position center.
    \item \textbf{Distance to Edge} ($d_{edge} = (S_t - c) / (u - c)$): The normalized displacement from the position center, where $u$ is the upper bound. Values of $\pm 1$ correspond to the range edges; $|d_{edge}| > 1$ indicates the price has exited the range.
    \item \textbf{Stein Signal} ($\theta$): The mean-reversion speed parameter of the OU process, truncated to $[0, 1]$.
    \item \textbf{Mean Deviation} ($\delta_\mu = (\mu - S_t)/S_t$): The normalized distance between the current price and the equilibrium price $\mu$ to which it is expected to revert.
    \item \textbf{Normalized Sigma} ($\tilde{\sigma}$): The diffusion parameter $\sigma$ normalized by price, clipped at $0.1$ to provide a stable volatility signal.
    \item \textbf{Active Fraction} ($\phi_{active}$): The cumulative fraction of time the position has spent in-range during the current episode.
    \item \textbf{Recent Volatility} ($v$): Rolling realized volatility (300-second window) clipped at $0.1$.
    \item \textbf{In-Range Indicator} ($1_\text{in}$): A binary flag indicating whether the current price $S_t$ is within the liquidity bounds.
\end{itemize}

For the RAmmStein-Width variant, the state is extended to 9 dimensions with the addition of a \textbf{Normalized Position Width} ($\tilde{w} = (w - w_{\min}) / (w_{\max} - w_{\min})$), where $w_{\min} = 1\%$ and $w_{\max} = 5\%$ are the minimum and maximum allowed widths. This enables the agent to condition its decisions on the current concentration level.

\subsection{Action Space}

The action space of the \rammstein\ agent is defined as a binary discrete set:
\begin{equation}
    \mathcal{A} = \{0, 1\}
\end{equation}
where the actions correspond to the following logic:

\begin{itemize}
    \item $a = 0$ (Hold): The agent remains inactive. The position center stays constant ($c_{t+1} = c_t$), and the system evolves according to the continuation regime.
    \item $a = 1$ (Rebalance): The agent triggers an immediate rebalancing event. The position center is updated to the current market price, $c_{t+1} = S_t$, and a fixed rebalancing cost $C$ is incurred.
\end{itemize}

\medskip{RAmmStein-Width Action Space}

RAmmStein-Width extends the action space to jointly optimize timing and position width:
\begin{equation}
    \mathcal{A}_W = \{0, 1, 2, 3, 4, 5\}
\end{equation}
where $a = 0$ is hold and $a \in \{1, \ldots, 5\}$ triggers a rebalance with width $w = a\%$. This introduces a width--fee tradeoff: narrower positions achieve higher concentration $\Lambda = 1/\sqrt{w}$ but exit the range faster, while wider positions sacrifice concentration for stability. The agent learns the regime-conditional optimal width.

\subsection{Reward Function}

The reward function encodes the agent's reward after subtracting operational expenses from the total fee revenue. More formally,

\begin{equation}
r_t = \underbrace{\frac{\Delta \text{Fees}_t - \Delta \text{Gas}_t}{K}}_{\text{Net PnL}} \cdot \lambda + \underbrace{\epsilon \cdot 1_{\text{in}}}_{\text{Active Bonus}}
\end{equation}

where:
\begin{itemize}
    \item $\Delta \text{Fees}_t$ Total revenue of the position during period $t$.
    \item $\Delta \text{Gas}_t$  Rebalancing costs incurred (zero if the action is 0).
    \item $K$  Initial capital put in the position. This is used to compute ROI.
    \item $\lambda$  A scaling factor to stabilize the training procedure.
    \item $\epsilon$  Introduces an inductive bias by encouraging the agent to be in the range.
\end{itemize}

This reward structure ensures that the agent directly optimizes net ROI.

For RAmmStein-Width, an additional concentration bonus encourages narrower positions when the agent is in-range:
\begin{equation}
r_t^W = r_t + \lambda_w \cdot \log(1/w) \cdot 1_{\text{in}}
\end{equation}
where $\lambda_w = 0.005$. This term rewards the agent for tighter concentration.

\subsection{Network Architecture}

The Q-function $Q(s, a)$ is approximated using a fully-connected neural network. For the base \rammstein\ agent, the input layer consists of 8 neurons corresponding to the state vector dimensions, followed by two hidden layers of 128 and 64 neurons with ReLU activations. The output layer comprises 2 neurons representing the Q-values for each action. For RAmmStein-Width, the input layer has 9 neurons (adding $\tilde{w}$) and the output layer has 6 neurons (one per action). To ensure training stability, a target network $Q^-$ is maintained as described in Section~II-F and synchronized with the online network ($\theta^- \leftarrow \theta$) every 100 training steps.

\subsection{Training Procedure}
The training procedure is detailed in Algorithm~\ref{alg:training}. We use the standard DDQN approach to mitigate overestimation bias by decoupling action selection from target value evaluation. The hyperparameters are listed in Table~\ref{tab:hyperparameters}.

\begin{algorithm}
\caption{\rammstein\ Training}
\label{alg:training}
\begin{algorithmic}[1]
\STATE Initialize Q-network $Q(s,a;\theta)$ with random weights
\STATE Initialize target network $Q^-$ with weights $\theta^- = \theta$
\STATE Initialize replay buffer $\mathcal{D}$ with capacity $N$
\FOR{episode $= 1$ to $M$}
    \STATE Sample random starting point in training data
    \STATE Initialize position at current price
    \FOR{$t = 1$ to $T$}
        \STATE Observe state $s_t$
        \STATE Select action $a_t = \begin{cases} \text{random} & \text{w.p. } \epsilon \\ \arg\max_a Q(s_t, a) & \text{otherwise} \end{cases}$
        \STATE Execute action, observe $r_t$, $s_{t+1}$
        \STATE Store $(s_t, a_t, r_t, s_{t+1})$ in $\mathcal{D}$
        \STATE Sample minibatch from $\mathcal{D}$
        \STATE Compute target: $y = r + \gamma Q^-(s', \arg\max_{a'} Q(s', a'))$
        \STATE Update $\theta$ by gradient descent on $(y - Q(s,a))^2$
        \IF{$t \mod 100 = 0$}
            \STATE $\theta^- \leftarrow \theta$
        \ENDIF
    \ENDFOR
    \STATE Decay $\epsilon$
\ENDFOR
\end{algorithmic}
\end{algorithm}

\begin{table}[h]
\centering
\caption{Training Hyperparameters}
\label{tab:hyperparameters}
\begin{tabular}{ll}
\toprule
\textbf{Parameter} & \textbf{Value} \\
\midrule
Learning rate & $10^{-4}$ \\
Discount factor $\gamma$ & 0.99 \\
Replay buffer size & 100,000 \\
Batch size & 128 \\
Target update frequency & 100 steps \\
$\epsilon$ start / end / decay & 1.0 / 0.05 / 0.9998 \\
Episodes & 150 \\
Episode length & 36,000 steps (10 hours) \\
\bottomrule
\end{tabular}
\end{table}

\section{Experimental Setup}

\subsection{Dataset and Preprocessing}

We utilize a comprehensive high-frequency dataset of ETH-USD trades observed on the Coinbase WebSocket feed, spanning two weeks from January 20 to February 3, 2026. This period captures significant market volatility, with a price range of \$2,108 to \$3,067 and a net price decline of 26.2\%. The raw data consists of 6.8 million individual trades and is aggregated into 1Hz OHLCV bars. The one-second resolution is critical for capturing mean-reversion behavior. Lower-frequency data, such as one-minute candles, would obscure the micro-level dynamics.

To get an unbiased estimate of the agent's performance and prevent look-ahead bias, we divide the dataset into training, validation, and testing subsets: the first 10 days (70\%) are used for training, the next 2 days (15\%) for validation and hyperparameter tuning, and the final 2 days (15\%) for out-of-sample testing.

\subsection{Environment and Fee Estimation}

The environment is configured to mimic a Uniswap V3-style pool with parameters listed in Table~\ref{tab:env_params}.
To estimate DEX trading volume from our CEX price source, we adopt a volume scaling approach: $V_{DEX} = \alpha \cdot V_{CEX}$, where $\alpha = 0.10$. This 10\% ratio is consistent with observed market structures for major pairs.

\begin{table}[h]
\centering
\caption{Environment Parameters}
\label{tab:env_params}
\begin{tabular}{ll}
\toprule
\textbf{Parameter} & \textbf{Value} \\
\midrule
Range width & 1\% (100 bps) \\
Pool fee tier ($\phi$) & 0.05\% \\
Gas cost ($G$) & \$2.00 USD \\
Initial capital ($K$) & \$10,000 USD \\
Pool TVL ($L_{pool}$) & \$10,000,000 USD \\
DEX/CEX volume ratio ($\alpha$) & 10\% \\
\bottomrule
\end{tabular}
\end{table}

The fee accrual per timestep is computed as $\text{Fee}_t = V_{DEX,t} \cdot \phi \cdot \frac{L_{LP} \cdot \lambda}{L_{pool}}$, where $\lambda$ is the concentration multiplier. The total rebalancing cost $C$ includes both the fixed gas cost $G$ and a proportional swap fee:
\begin{equation}
C = \underbrace{\phi \cdot 0.5 \cdot K}_{\text{Swap fee}} + \underbrace{G}_{\text{Gas}}
\end{equation}
This assumes approximately 50\% of the position is swapped to maintain the desired inventory delta during recentering.

\subsection{OU Parameter Estimation}

The OU parameters are estimated via rolling Ordinary Least Squares (OLS) on a 1,800-second window. We regress $S_{t+1} - S_t$ on $S_t$ to derive the coefficients $(\hat{\alpha}_t, \hat{\beta}_t)$, from which the OU parameters are extracted:
\begin{equation}
\hat{\theta}_t = \frac{-\hat{\beta}_t}{\Delta t}, \quad \hat{\mu}_t = -\frac{\hat{\alpha}_t}{\hat{\beta}_t}
\end{equation}

In the dataset, we observe strong mean-reversion ($\theta > 0.01$) in 8.7\% of observations. The vast majority exhibit weak mean-reversion. The median $\theta$ value is $0.0056$, corresponding to a half-life of slightly more than two minutes. This validates our hypothesis that deviations arising from market microstructure typically revert within minutes. During such periods, the optimal strategy for the agent is to wait rather than react.

\section{Comparative Strategies}

To evaluate \rammstein, we compare its performance against four baseline strategies and the RAmmStein-Width variant.

\subsection{Set \& Forget (Omniscient Oracle)}
This strategy represents an upper bound on passive performance. The liquidity position is initialized with a range $[S_{\text{min}}, S_{\text{max}}]$ so that the position is active during the entire test.
\begin{equation}
c = \frac{S_{\text{min}} + S_{\text{max}}}{2}, \quad w = \frac{S_{\text{max}} - S_{\text{min}}}{2c}
\end{equation}
This strategy is unrealistic as it requires future knowledge. However, it achieves 100\% active time and has no additional costs after initializing the position.

\subsection{Fixed Passive}
A 1\% range is set at the initial price and never adjusted ($c = S_0, w = 0.01$). If the price exits the range, the position remains inactive until it returns naturally. This strategy aims to completely avoid rebalancing costs.

\subsection{Fixed Active (Greedy Rebalancing)}
In this strategy, the position is always in a narrow range around the current price. If the price exits the range, the position is immediately recentered.
\begin{equation}
\text{If } S_t \notin [c(1-w), c(1+w)] \implies c \leftarrow S_t
\end{equation}
This strategy aims to maximize the active time and at the same time incurs the maximum rebalancing costs. We show that this myopic behavior destroys capital through excessive friction costs.

\subsection{LSTM Predictor}
In this strategy, a Long Short-Term Memory (LSTM) network is trained to predict the future price $S_{t+H}$ based on historical prices, volume, and OU parameters.
\begin{equation}
\hat{S}_{t+H} = \text{LSTM}(S_{t-L:t}, V_{t-L:t}, \theta_{t-L:t}, \mu_{t-L:t}, \sigma_{t-L:t})
\end{equation}
This strategy recenters the liquidity position at the predicted price. It is more sophisticated than simple threshold-based approaches, but critically, it does not account for rebalancing costs.

\subsection{\rammstein\ (Proposed)}
Our DDQN-based agent observes the full eight-dimensional state vector and selects the optimal action: $a_t = \arg\max_a Q(s_t, a; \theta^*)$. The agent learns to rebalance only when the price has exited the range and $\theta$ is low, indicating that a natural return is unlikely.

\subsection{RAmmStein-Width (Proposed)}
This variant extends \rammstein\ to jointly optimize rebalancing timing and position width. The agent observes a 9-dimensional state (adding normalized width $\tilde{w}$) and selects from 6 actions: hold or rebalance with width $w \in \{1\%, 2\%, 3\%, 4\%, 5\%\}$. This tests whether the DDQN framework generalizes to richer action spaces and whether the agent can learn regime-conditional width selection.

\section{Evaluation Metrics}

We use the following metrics to compare the different strategies:
\begin{itemize}
    \item \textbf{Active Fraction ($\Phi$)}: The percentage of time the market price resides within the position's range. Higher values of this metric represent higher asset utilization.
    \item \textbf{Normalized Liquidity ($\Lambda$)}: A measure of the concentration of the position. This depends inversely on the $\sqrt{w}$, ($1/\sqrt{w}$). A 1\% width achieves $\Lambda \approx 10\times$.
    \item \textbf{Rebalance Count ($N$)}: The total number of discrete rebalancing actions. Lower counts represent lower operational costs.
    \item \textbf{Net ROI}: The primary metric that the strategies aim to optimize. It is the return on investment after all operational costs:
    \begin{equation}
    \text{Net ROI} = \frac{\text{Total Fees} - \text{Total Gas}}{\text{Initial Capital}}
    \end{equation}
\end{itemize}

\section{Experimental Results}

In this section, we present the results of evaluating the different strategies on the out-of-sample subset of the dataset.

\subsection{Main Comparison}

The primary results are summarized in Table~\ref{tab:main_results}. \rammstein\ achieves the highest net ROI of 1.60\% among all realistic (non-omniscient) strategies. The omniscient Set \& Forget baseline achieves 1.94\% but requires future knowledge of the price range.

\begin{table}[h]
\centering
\caption{Strategy Comparison on Test Data (\$10M TVL, 1\% width, \$2 gas)}
\label{tab:main_results}
\begin{tabular}{lrrrrr}
\toprule
\textbf{Strategy} & \textbf{Active\%} & \textbf{Rebal.} & \textbf{Fees (\$)} & \textbf{Gas (\$)} & \textbf{Net ROI} \\
\midrule
Set \& Forget (Oracle) & 100.0\% & 1 & 198 & 4.5 & 1.94\% \\
Fixed Passive & 19.7\% & 1 & 117 & 4.5 & 1.13\% \\
Fixed Active (Greedy) & 100.0\% & 344 & 644 & 1,483 & $-$8.40\% \\
LSTM Predictor & 95.1\% & 237 & 580 & 1,036 & $-$4.55\% \\
\textbf{\rammstein} & \textbf{57.7\%} & \textbf{51} & \textbf{389} & \textbf{228} & \textbf{1.60\%} \\
RAmmStein-Width & 9.3\% & 9 & 75 & 40 & 0.35\% \\
\bottomrule
\end{tabular}
\end{table}

The results clearly illustrate the \textbf{LP Rebalancing Paradox}: Fixed Active earns the highest gross fees (\$644) yet suffers the worst net ROI ($-8.4\%$) because its 344 rebalances incur \$1,483 in gas costs. \rammstein\ achieves an 85\% reduction in rebalancing frequency (51 vs.\ 344) while maintaining sufficient active time to earn \$389 in fees, converting this to a positive 1.60\% net return, a 10 percentage point improvement over the greedy strategy.

RAmmStein-Width agent takes a different approach: the agent discovers extreme parsimony on its own, executing only 9 rebalances with \$40 in gas. While its 0.35\% ROI is lower than \rammstein, it demonstrates that the DDQN framework generalizes to richer action spaces and produces the lowest transaction cost of any active strategy.

\subsection{The $\theta$ Effect}

\begin{figure}[h]
\centering
\includegraphics[width=0.9\columnwidth]{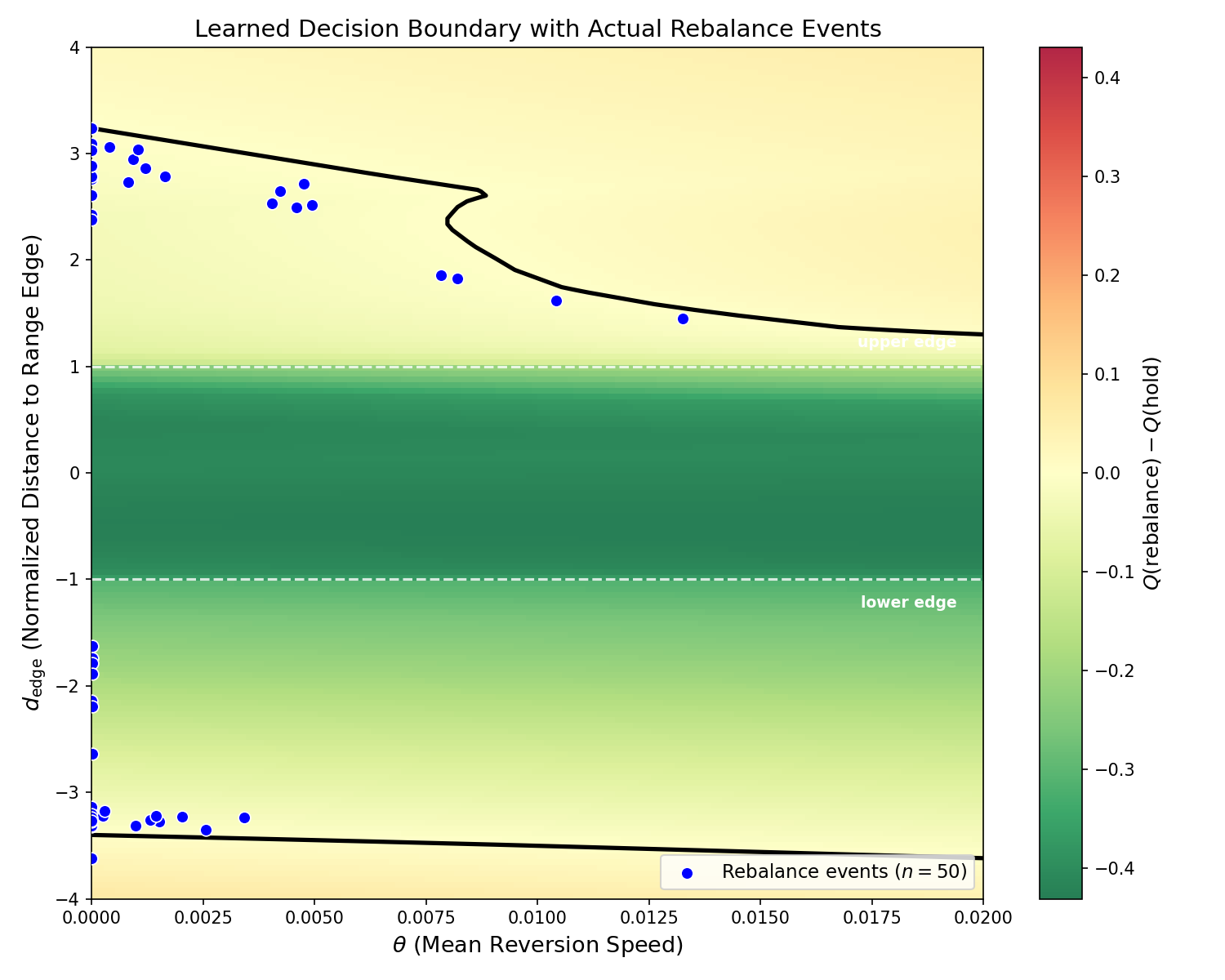}
\caption{Learned decision boundary with actual rebalance events (blue dots, $n=50$). The heatmap shows $Q(\text{rebalance}) - Q(\text{hold})$; green = hold preferred, red = rebalance preferred. White dashed lines mark the range edges ($d_{\text{edge}} = \pm 1$). All rebalances occur outside the range at low $\theta$.}
\label{fig:decision_boundary}
\end{figure}

To understand the learned policy, we analyze the agent's $Q$-values across the $(\theta, d_{\text{edge}})$ state space and overlay the actual rebalance events from the backtest (Figure~\ref{fig:decision_boundary}). The heatmap shows $Q(\text{rebalance}) - Q(\text{hold})$; the black contour marks the decision boundary where the agent switches from hold (green) to rebalance (red). White dashed lines indicate the range edges ($d_{\text{edge}} = \pm 1$); values beyond $\pm 1$ correspond to prices that have exited the range.

The overlay reveals three key properties of the learned policy:
\begin{itemize}
    \item \textbf{Out-of-range rebalancing}: 50 of 51 rebalances occur at $|d_{\text{edge}}| > 1$, i.e., after the price has already exited the range. The entire in-range region is green (hold), confirming that the agent never rebalances while earning fees. This is consistent with the impulse control formulation: intervention is only optimal once the position has become inactive.
    \item \textbf{$\theta$-dependent threshold}: The decision boundary drops as $\theta$ increases: at $\theta \approx 0$, the agent requires $d_{\text{edge}} > 3$ before rebalancing, while at $\theta \approx 0.02$ the threshold falls to $d_{\text{edge}} \approx 1.3$. The agent has learned that rebalancing is more \emph{durable} in mean-reverting conditions: the price tends to remain in the new range, so the gas cost is justified at smaller excursions. In trending conditions ($\theta \approx 0$), any rebalance is quickly undone, so the agent demands a much larger displacement before spending gas.
    \item \textbf{Rebalances cluster at low $\theta$, large $|d_{\text{edge}}|$}: Despite the higher threshold at low $\theta$, trending prices produce excursions large enough to cross it. At high $\theta$, the price mean-reverts before reaching even the lower threshold. The net effect is that all observed rebalances occur at $\theta < 0.015$ (median $\approx 0$) with $|d_{\text{edge}}|$ ranging from 1.5 to 3.6.
    \item \textbf{Asymmetric rebalancing}: On the negative side, rebalances span a range of excursions ($d_{\text{edge}} \in [-3.6, -1.5]$), while on the positive side they are concentrated at larger values ($d_{\text{edge}} \in [1.5, 3.2]$). Several negative-side rebalances fall inside the hold region of this 2D projection, indicating that the remaining state dimensions ($\delta_\mu$, $\sigma$, volatility) contributed to the decision. A complete characterization of this asymmetry requires analysis beyond the $(\theta, d_{\text{edge}})$ slice.
\end{itemize}

This analysis confirms that \rammstein\ has internalized the Stein Signal: the agent uses $\theta$ to modulate the rebalancing threshold, setting it high in trending conditions (where rebalances are wasteful) and low in mean-reverting conditions (where they are durable), achieving the 85\% reduction in rebalancing frequency observed in the main results.

\subsection{Gas Sensitivity Analysis}

We also evaluate the strategies across varying gas cost regimes in Table~\ref{tab:gas_sensitivity}. The three active strategies exhibit strikingly different gas resilience profiles.

\begin{table}[h]
\centering
\caption{Net ROI vs.\ Gas Cost}
\label{tab:gas_sensitivity}
\begin{tabular}{lrrr}
\toprule
\textbf{Gas (USD)} & \textbf{Fixed Active} & \textbf{RAmmStein} & \textbf{RAmmStein-Width} \\
\midrule
\$1.00 & $-$4.96\% & 1.86\% & 0.44\% \\
\$2.00 & $-$8.40\% & 1.60\% & 0.35\% \\
\$5.00 & $-$18.72\% & 0.83\% & 0.08\% \\
\$10.00 & $-$35.94\% & $-$0.50\% & $-$0.36\% \\
\$20.00 & $-$70.38\% & $-$3.16\% & $-$1.24\% \\
\bottomrule
\end{tabular}
\end{table}

Fixed Active is unprofitable at \emph{all} tested gas levels due to its 344 rebalances. \rammstein\ remains profitable up to approximately \$7 gas and dominates RAmmStein-Width at every profitable gas level. At \$10+ gas both strategies are unprofitable, but RAmmStein-Width ($-0.36\%$) degrades more slowly than \rammstein\ ($-0.50\%$) due to its 9 vs.\ 51 rebalances. This crossover illustrates that extreme parsimony provides a floor on losses during high-gas regimes, even though it sacrifices upside at lower gas costs.

\subsection{Width Selection Analysis}

\begin{figure}[h]
\centering
\includegraphics[width=\columnwidth]{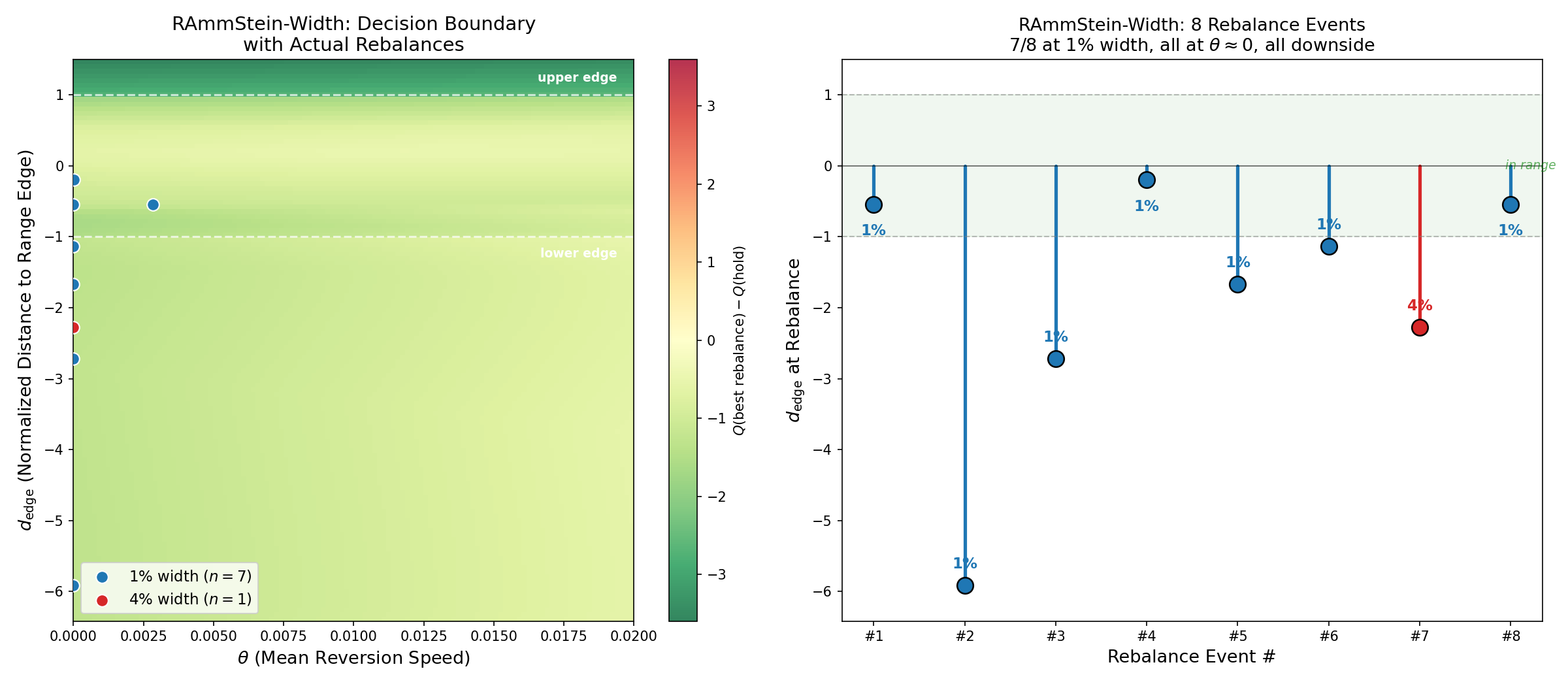}
\caption{RAmmStein-Width: $Q(\text{best rebalance}) - Q(\text{hold})$ landscape with actual rebalance events (left), and per-event $d_{\text{edge}}$ with chosen width annotation (right). All 8 rebalances are downside exits at $\theta \approx 0$; 7 of 8 select 1\% width.}
\label{fig:width_distribution}
\end{figure}

Figure~\ref{fig:width_distribution} overlays the 8 actual rebalance events of RAmmStein-Width on its $Q$-value landscape (left) and shows each event's $d_{\text{edge}}$ and chosen width (right). Three behaviors emerge: (1) the $(\theta, d_{\text{edge}})$ projection is almost entirely hold-preferred (green), indicating that the rebalancing decision is driven by the remaining state dimensions ($\delta_\mu$, $\sigma$, width, volatility) rather than by $\theta$ and $d_{\text{edge}}$ alone; (2) all 8 rebalances occur on the \emph{downside} ($d_{\text{edge}} < 0$), with none on upward price exits; and (3) the agent overwhelmingly selects the tightest width (1\%) in 7 of 8 events, maximizing fee concentration. This indicates that the agent has learned extreme selectivity about \emph{when} to act, but is aggressive about concentration \emph{when} it does.

\subsection{Fee and Gas Decomposition}

Table~\ref{tab:fee_gas_decomp} decomposes the returns of each strategy into fee revenue, gas expenditure, and the resulting fee-to-gas ratio. The decomposition reveals the core mechanism behind the LP Rebalancing Paradox: Fixed Active earns the most fees (\$644) but spends 2.3$\times$ more on gas (\$1,483), yielding a fee-to-gas ratio of only 0.4. In contrast, \rammstein\ achieves a ratio of 1.7 and RAmmStein-Width achieves 1.9, the highest among active strategies.

\begin{table}[h]
\centering
\caption{Fee and Gas Decomposition}
\label{tab:fee_gas_decomp}
\begin{tabular}{lrrrr}
\toprule
\textbf{Strategy} & \textbf{Fees (\$)} & \textbf{Gas (\$)} & \textbf{Net PnL (\$)} & \textbf{Fee/Gas} \\
\midrule
Set \& Forget (Oracle) & 198 & 4.5 & 194 & 44.1 \\
Fixed Passive & 117 & 4.5 & 113 & 26.1 \\
Fixed Active & 644 & 1,483 & $-$840 & 0.4 \\
LSTM Predictor & 580 & 1,036 & $-$455 & 0.6 \\
\textbf{\rammstein} & \textbf{389} & \textbf{228} & \textbf{160} & \textbf{1.7} \\
RAmmStein-Width & 75 & 40 & 35 & 1.9 \\
\bottomrule
\end{tabular}
\end{table}

\subsection{Decision Analysis}

The experimental results show that while Fixed Active treats every range exit as a reason to rebalance, \rammstein\ learned to defer rebalancing when $\theta$ is elevated, waiting for the price to revert naturally. Analysis of the 51 rebalance events confirms that all occur outside the range at very low $\theta$ (median $\approx 0$), i.e., only when a natural return is unlikely. This selective approach resulted in an 85\% reduction in rebalances and a 10 percentage point ROI improvement over the greedy strategy. RAmmStein-Width pushes this further, learning to use wider positions that require even fewer interventions.

\section{Discussion}

\subsection{The Value of Regime Awareness}

Our results clearly show that incorporating market regime information significantly improves LP returns. The 10 percentage point gap between Fixed Active ($-8.4\%$) and \rammstein\ ($+1.6\%$) demonstrates the economic magnitude of regime-aware decision making. The OU parameters provide well-defined features capturing regime dynamics that are invisible to models based on price information alone. Our agent learns to quantify the value of the waiting option.

\subsection{Dynamic Width as an Optimization Axis}

RAmmStein-Width demonstrates that position width is a learnable dimension alongside rebalancing timing. The agent's discovery of extreme parsimony (9 rebalances, 9.3\% active time) was not hardcoded but emerged from the reward structure. This suggests that the concentration--stability tradeoff is a rich optimization surface that warrants further exploration with continuous action spaces.

\subsection{Implications for LP Tooling}

Existing tools for LP management typically follow a greedy strategy that chases active time over cost efficiency. Our analysis suggests that ROI can be improved by moving away from fixed heuristics towards models that base decisions on:

\begin{itemize}
    \item OU parameters that measure mean-reversion strength.
    \item Conditions for triggering rebalance.
    \item Costs for rebalancing events.
    \item Dynamic width selection conditioned on market regime.
\end{itemize}

\subsection{Limitations and Future Work}

Several limitations warrant acknowledgment. Our Net ROI metric captures fee revenue minus rebalancing costs but does not account for impermanent loss or portfolio revaluation; the position capital is held constant between rebalances. Our work focuses on a single asset pair; measuring the effectiveness of the approach on more volatile and lower-liquidity pairs remains to be validated. We also assume a fixed pool share for the liquidity provider and that the OU parameters are locally stationary within the estimation window.

Extensions of our work include: (1) continuous width optimization via policy gradient methods (e.g., SAC), which would remove the discretization of the width action space; (2) joint center placement towards the OU mean $\mu$, allowing anticipatory repositioning; (3) multi-pool allocation across correlated pairs; (4) integration of gas price prediction into the state to further optimize intervention timing; and (5) preventing or minimizing MEV risks during rebalancing.

\section{Conclusion}

This paper introduced RAmmStein, a Deep Reinforcement Learning framework for managing concentrated liquidity positions in AMMs. By framing LP rebalancing as an impulse control problem and incorporating the Ornstein-Uhlenbeck parameters of price dynamics as a prior, our agent learns a dynamic laziness boundary that adapts to the market conditions.

Extensive backtesting on 1Hz data with realistic pool parameters demonstrated that \rammstein\ achieves the highest net ROI (1.60\%) among all realistic active strategies, while greedy strategies destroy 8.4\% of capital through excessive rebalancing. The RAmmStein-Width extension further demonstrates the framework's extensibility, learning extreme parsimony and degrading more gracefully under elevated gas costs.

The key takeaway from our results is that deferring rebalancing decisions preserves capital that would otherwise be eroded by friction costs. Greedy strategies that chase 100\% active time convert fee revenue into gas expenditure, while regime-aware laziness converts the same market conditions into positive returns.

As DeFi matures, we anticipate that ML approaches that incorporate signals from quantitative finance will be increasingly used in liquidity management. \rammstein\ presents a step towards that automation.

\section*{Note}
An earlier version of this paper was submitted to the Designing DeFi workshop (\url{https://www.designingdefi.xyz/}).

\bibliographystyle{IEEEtran}
\bibliography{rammstein}

\begin{thebibliography}{10}
\providecommand{\url}[1]{#1}
\csname url@samestyle\endcsname
\providecommand{\newblock}{\relax}
\providecommand{\bibinfo}[2]{#2}
\providecommand{\BIBentrySTDinterwordspacing}{\spaceskip=0pt\relax}
\providecommand{\BIBentryALTinterwordstretchfactor}{4}
\providecommand{\BIBentryALTinterwordspacing}{\spaceskip=\fontdimen2\font plus
\BIBentryALTinterwordstretchfactor\fontdimen3\font minus
  \fontdimen4\font\relax}
\providecommand{\BIBforeignlanguage}[2]{{%
\expandafter\ifx\csname l@#1\endcsname\relax
\typeout{** WARNING: IEEEtran.bst: No hyphenation pattern has been}%
\typeout{** loaded for the language `#1'. Using the pattern for}%
\typeout{** the default language instead.}%
\else
\language=\csname l@#1\endcsname
\fi
#2}}
\providecommand{\BIBdecl}{\relax}
\BIBdecl

\bibitem{adams2021uniswap}
\BIBentryALTinterwordspacing
H.~Adams, N.~Zinsmeister, and D.~Robinson, ``Uniswap v2 core,'' 2020. [Online].
  Available: \url{https://app.uniswap.org/whitepaper.pdf}
\BIBentrySTDinterwordspacing

\bibitem{adams2021uniswapv3}
\BIBentryALTinterwordspacing
H.~Adams, N.~Zinsmeister, M.~Salem, R.~Keefer, and D.~Robinson, ``Uniswap v3
  core,'' 2021. [Online]. Available:
  \url{https://app.uniswap.org/whitepaper-v3.pdf}
\BIBentrySTDinterwordspacing

\bibitem{egorov2019stableswap}
\BIBentryALTinterwordspacing
M.~Egorov, ``Stableswap -- efficient mechanism for stablecoin liquidity,''
  2019. [Online]. Available:
  \url{https://berkeley-defi.github.io/assets/material/StableSwap.pdf}
\BIBentrySTDinterwordspacing

\bibitem{orbital2024}
\BIBentryALTinterwordspacing
D.~White, D.~Robinson, and C.~Moallemi, ``Orbital,'' Paradigm, 2025. [Online].
  Available: \url{https://www.paradigm.xyz/2025/06/orbital}
\BIBentrySTDinterwordspacing

\bibitem{elsts2021uniswapv3math}
A.~Elsts, ``Liquidity math in uniswap v3,'' \emph{SSRN Electronic Journal},
  2023.

\bibitem{loesch2021impermanent}
\BIBentryALTinterwordspacing
S.~Loesch, N.~Hindman, M.~B. Richardson, and N.~Welch, ``Impermanent loss in
  uniswap v3,'' 2021. [Online]. Available:
  \url{https://arxiv.org/abs/2111.09192}
\BIBentrySTDinterwordspacing

\bibitem{lambert2022uniswap}
G.~Lambert, ``Uniswap v3: A quant framework,'' Lambert's Newsletter, 2022.

\bibitem{milionis2022automated}
\BIBentryALTinterwordspacing
J.~Milionis, C.~C. Moallemi, T.~Roughgarden, and A.~L. Zhang, ``Automated
  market making and loss-versus-rebalancing,'' 2022. [Online]. Available:
  \url{https://arxiv.org/abs/2208.06046}
\BIBentrySTDinterwordspacing

\bibitem{heimbach2021behavior}
L.~Heimbach, Y.~Wang, and R.~Wattenhofer, ``Behavior of liquidity providers in
  decentralized exchanges,'' in \emph{Proc. Crypto Valley Conference on
  Blockchain Technology (CVCBT)}, 2021.

\bibitem{berg2021empirical}
\BIBentryALTinterwordspacing
J.~A. Berg, R.~Fritsch, L.~Heimbach, and R.~Wattenhofer, ``An empirical study
  of market inefficiencies in uniswap and sushiswap,'' in \emph{Proc.
  Blockchain}, 2022. [Online]. Available:
  \url{https://arxiv.org/abs/2203.07774}
\BIBentrySTDinterwordspacing

\bibitem{heimbach2022risks}
L.~Heimbach, E.~Schertenleib, and R.~Wattenhofer, ``Risks and returns of
  uniswap v3 liquidity providers,'' in \emph{Proc. 4th ACM Conference on
  Advances in Financial Technologies (AFT)}, 2022.

\bibitem{fritsch2021concentrated}
R.~Fritsch, ``Concentrated liquidity in automated market makers,'' in
  \emph{Proc. ACM CCS Workshop on Decentralized Finance and Security (DeFi)},
  2021.

\bibitem{evans2020liquidity}
\BIBentryALTinterwordspacing
A.~Evans, ``Liquidity provider returns in geometric mean markets,'' 2020.
  [Online]. Available: \url{https://arxiv.org/abs/2006.08806}
\BIBentrySTDinterwordspacing

\bibitem{cartea2024predictable}
A.~Cartea, F.~Drissi, and M.~Monga, ``Decentralized finance and automated
  market making: Predictable loss and optimal liquidity provision,'' \emph{SIAM
  Journal on Financial Mathematics}, vol.~15, no.~3, pp. 931--959, 2024.

\bibitem{milionis2023myersonian}
J.~Milionis, C.~C. Moallemi, and T.~Roughgarden, ``A myersonian framework for
  optimal liquidity provision in automated market makers,'' 2023.

\bibitem{baron2023online}
Y.~Bar-On and Y.~Mansour, ``Uniswap liquidity provision: An online learning
  approach,'' in \emph{Lecture Notes in Computer Science}.\hskip 1em plus 0.5em
  minus 0.4em\relax Springer, 2023, pp. 247--261.

\bibitem{fan2024strategic}
\BIBentryALTinterwordspacing
Z.~Fan, F.~Marmolejo-Coss\'{i}o, D.~J. Moroz, M.~Neuder, R.~Rao, and D.~C.
  Parkes, ``Strategic liquidity provision in uniswap v3,'' 2023. [Online].
  Available: \url{https://arxiv.org/abs/2106.12033}
\BIBentrySTDinterwordspacing

\bibitem{urusov2025tau}
\BIBentryALTinterwordspacing
A.~Urusov, R.~Berezovskiy, A.~Krestenko, and A.~Kornilov, ``Liquidity provision
  with $\tau$-reset strategies: a dynamic historical liquidity approach,''
  2025. [Online]. Available: \url{https://arxiv.org/abs/2505.15338}
\BIBentrySTDinterwordspacing

\bibitem{zhang2023adaptive}
\BIBentryALTinterwordspacing
H.~Zhang, X.~Chen, and L.~F. Yang, ``Adaptive liquidity provision in uniswap v3
  with deep reinforcement learning,'' 2023. [Online]. Available:
  \url{https://arxiv.org/abs/2309.10129}
\BIBentrySTDinterwordspacing

\bibitem{xu2025improving}
\BIBentryALTinterwordspacing
H.~Xu and A.~Brini, ``Improving defi accessibility through efficient liquidity
  provisioning with deep reinforcement learning,'' in \emph{Proc. AAAI}, 2025.
  [Online]. Available: \url{https://arxiv.org/abs/2501.07508}
\BIBentrySTDinterwordspacing

\bibitem{jaimungal2023optimal}
\BIBentryALTinterwordspacing
S.~Jaimungal, Y.~F. Saporito, M.~O. Souza, and Y.~Thamsten, ``Optimal trading
  in automatic market makers with deep learning,'' 2023. [Online]. Available:
  \url{https://arxiv.org/abs/2304.02180}
\BIBentrySTDinterwordspacing

\bibitem{bensoussan1984impulse}
A.~Bensoussan and J.-L. Lions, \emph{Impulse Control and Quasi-variational
  Inequalities}.\hskip 1em plus 0.5em minus 0.4em\relax Gauthier-Villars, 1984.

\bibitem{uhlenbeck1930theory}
G.~E. Uhlenbeck and L.~S. Ornstein, ``On the theory of the brownian motion,''
  \emph{Physical Review}, vol.~36, no.~5, pp. 823--841, 1930.

\bibitem{vasicek1977equilibrium}
O.~Vasicek, ``An equilibrium characterization of the term structure,''
  \emph{Journal of Financial Economics}, vol.~5, no.~2, pp. 177--188, 1977.

\bibitem{efron1977stein}
B.~Efron and C.~Morris, ``Stein's paradox in statistics,'' \emph{Scientific
  American}, vol. 236, no.~5, pp. 119--127, 1977.

\bibitem{silver2016mastering}
D.~Silver, A.~Huang, C.~J. Maddison, A.~Guez, L.~Sifre, G.~van~den Driessche,
  J.~Schrittwieser, I.~Antonoglou, V.~Panneershelvam, M.~Lanctot, S.~Dieleman,
  D.~Grewe, J.~Nham, N.~Kalchbrenner, I.~Sutskever, T.~Lillicrap, M.~Leach,
  K.~Kavukcuoglu, T.~Graepel, and D.~Hassabis, ``Mastering the game of go with
  deep neural networks and tree search,'' \emph{Nature}, vol. 529, no. 7587,
  pp. 484--489, 2016.

\bibitem{levine2016end}
S.~Levine, C.~Finn, T.~Darrell, and P.~Abbeel, ``End-to-end training of deep
  visuomotor policies,'' \emph{Journal of Machine Learning Research}, vol.~17,
  no.~39, pp. 1--40, 2016.

\bibitem{jiang2017deep}
\BIBentryALTinterwordspacing
Z.~Jiang, D.~Xu, and J.~Liang, ``A deep reinforcement learning framework for
  the financial portfolio management problem,'' 2017. [Online]. Available:
  \url{https://arxiv.org/abs/1706.10059}
\BIBentrySTDinterwordspacing

\bibitem{spooner2018market}
T.~Spooner, J.~Fearnley, R.~Savani, and A.~Koukorinis, ``Market making via
  reinforcement learning,'' in \emph{Proceedings of AAMAS}, 2018.

\bibitem{ning2021double}
B.~Ning, F.~H.~T. Ling, and S.~Jaimungal, ``Double deep q-learning for optimal
  execution,'' \emph{Applied Mathematical Finance}, vol.~28, no.~4, pp.
  361--380, 2021.

\bibitem{mamageishvili2023batch}
\BIBentryALTinterwordspacing
A.~Mamageishvili and E.~Felten, ``Efficient rollup batch posting strategy on
  base layer,'' 2023. [Online]. Available:
  \url{https://fc23.ifca.ai/wtsc/WTSC23_1.pdf}
\BIBentrySTDinterwordspacing

\bibitem{sutton2018reinforcement}
R.~S. Sutton and A.~G. Barto, \emph{Reinforcement Learning: An Introduction},
  2nd~ed.\hskip 1em plus 0.5em minus 0.4em\relax MIT Press, 2018.

\bibitem{mnih2015human}
V.~Mnih, K.~Kavukcuoglu, D.~Silver, A.~A. Rusu, J.~Veness, M.~G. Bellemare,
  A.~Graves, M.~Riedmiller, A.~K. Fidjeland, G.~Ostrovski, S.~Petersen,
  C.~Beattie, A.~Sadik, I.~Antonoglou, H.~King, D.~Kumaran, D.~Wierstra,
  S.~Legg, and D.~Hassabis, ``Human-level control through deep reinforcement
  learning,'' \emph{Nature}, vol. 518, no. 7540, pp. 529--533, 2015.

\bibitem{van2016deep}
H.~Van~Hasselt, A.~Guez, and D.~Silver, ``Deep reinforcement learning with
  double q-learning,'' in \emph{Proceedings of AAAI}, 2016.

\bibitem{elliott2005}
R.~J. Elliott, J.~Van Der~Hoek, and W.~P. Malcolm, ``Pairs trading,''
  \emph{Quantitative Finance}, vol.~5, no.~3, pp. 271--276, 2005.

\end{thebibliography}

\end{document}